\documentclass[10pt,twocolumn,letterpaper]{article}

\usepackage[pagenumbers]{cvpr} 
\definecolor{cvprblue}{rgb}{0.21,0.49,0.74}
\usepackage[pagebackref,breaklinks,colorlinks,allcolors=cvprblue]{hyperref}

\usepackage[ruled,linesnumbered]{algorithm2e}

\def\paperID{**} 
\def\confName{CVPR}
\def\confYear{2026}

\title{Task-Oriented Data Synthesis and Control-Rectify Sampling for Remote Sensing Semantic Segmentation}

\author{Yunkai Yang$^{1}$, 
Yudong Zhang$^{2}$,
Kunquan Zhang$^{1}$,
Jinxiao Zhang$^{2}$,
Xinying Chen$^{3}$,\\
Haohuan Fu$^{2}$,
Runmin Dong$^{1}$\thanks{Corresponding author}
\\
$^1$Sun Yat-Sen University ~~~\\
$^2$Tsinghua University~~~
$^3$Beijing Institute of Technology~~~\\
{\tt\small yangyk26@mail2.sysu.edu.cn, dongrm3@mail.sysu.edu.cn}
}

\begin{document}

\maketitle
\begin{abstract}
With the rapid progress of controllable generation, training data synthesis has become a promising way to expand labeled datasets and alleviate manual annotation in remote sensing (RS). However, the complexity of semantic mask control and the uncertainty of sampling quality often limit the utility of synthetic data in downstream semantic segmentation tasks. To address these challenges, we propose a task-oriented data synthesis framework (TODSynth), including a Multimodal Diffusion Transformer (MM-DiT) with unified triple attention and a plug-and-play sampling strategy guided by task feedback. Built upon the powerful DiT-based generative foundation model, we systematically evaluate different control schemes, showing that a text–image–mask joint attention scheme combined with full fine-tuning of the image and mask branches significantly enhances the effectiveness of RS semantic segmentation data synthesis, particularly in few-shot and complex-scene scenarios. Furthermore, we propose a control-rectify flow matching (CRFM) method, which dynamically adjusts sampling directions guided by semantic loss during the early high-plasticity stage, mitigating the instability of generated images and bridging the gap between synthetic data and downstream segmentation tasks. Extensive experiments demonstrate that our approach consistently outperforms state-of-the-art controllable generation methods, producing more stable and task-oriented synthetic data for RS semantic segmentation. The code is available at \url{https://github.com/Yunkai-Yang/crfm}.
\end{abstract}    
\section{Introduction}
\label{sec:intro}

\begin{figure*}[t]
 \centering
 \includegraphics[width=1.0\linewidth,height=0.6\linewidth,keepaspectratio]{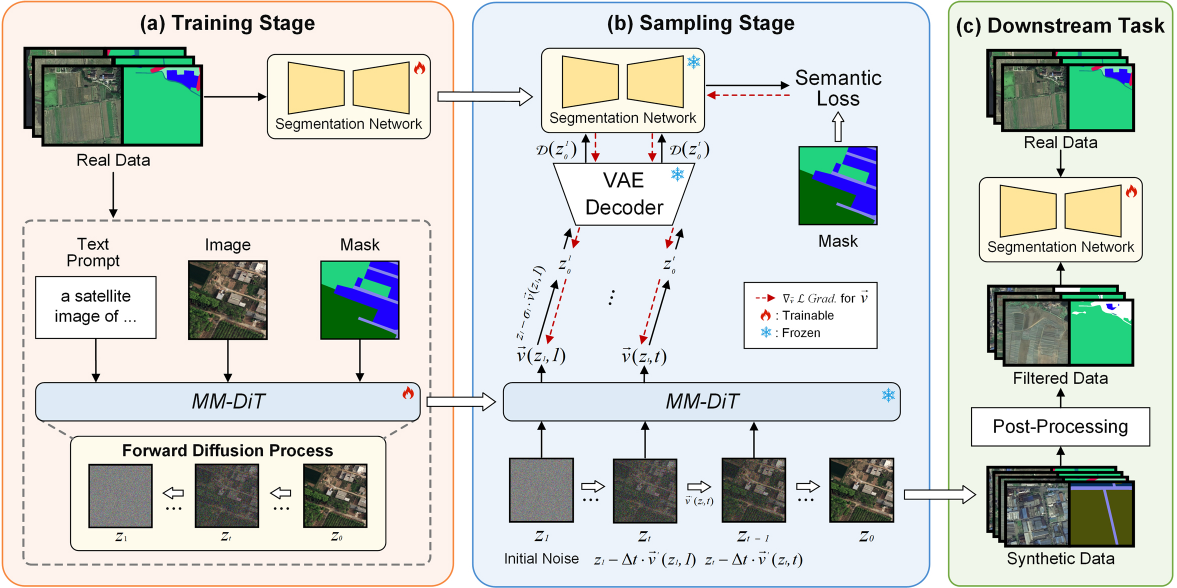}
 \caption{The overall workflow of the task-oriented data synthesis framework (TODSynth) consists of three stages: (a) Training stage using an MM-DiT generative model conditioned on text and mask. (b) Sampling stage with the proposed control-rectify flow matching (CRFM). (c) Downstream tasks trained on a combination of real and synthetic data.}
 \label{fig:1}
\vspace{-2mm}
\end{figure*}

Semantic segmentation in remote sensing (RS) imagery plays a crucial role in land-use classification~\cite{11134495}, environmental monitoring~\cite{chen2022contrasting}, and thematic mapping~\cite{li2025segearth}. However, compared with natural image segmentation, RS semantic segmentation faces substantial challenges due to the complex diversity of land-cover types across spatial resolutions and regions, the difficulty in collecting samples of rare classes, and the requirement for domain-specific expertise. Building large-scale pixel-wise labeled datasets for RS tasks is therefore time-consuming and expensive~\cite{dong2023large}. Recently, with the rapid advancement of generative models, data synthesis has emerged as a promising alternative to alleviate annotation scarcity and enrich training data for downstream segmentation models~\cite{DONG2026103839,namekata2024emerdiff}.

Early progress in text-to-image diffusion models~\cite{gao2024luminat2xtransformingtextmodality,li2024hunyuan,peebles2023scalable,podell2023sdxl,esser2024scaling} demonstrated remarkable capabilities in generating diverse and realistic samples for classification tasks. Controllable generation has further extended this paradigm to layout-to-image for object detection\cite{tang2025aerogen, ye2025object} and mask-to-image for semantic segmentation~\cite{pan2025earthsynth, tang2025terragen}, as exemplified by ControlNet~\cite{zhang2023adding} and its variants~\cite{li2024controlnet++, zhao2023uni, peng2024controlnext}. The emergence of transformer-based architectures~\cite{esser2024scaling,li2024hunyuan,peebles2023scalable} such as Diffusion Transformers (DiTs) has significantly improved generation quality. These powerful pre-trained generation models, such as SDXL~\cite{podell2023sdxl} and Stable Diffusion 3.5~\cite{stabilityai_sd35_release_2024}, provide a strong foundation for synthetic data generation and have been widely adapted for natural scene image synthesis.

Despite the maturity of ControlNet-style conditioning mechanisms, effectively controlling DiT-based generative models remains an open challenge~\cite{duan2025dit4sr}. 
While some studies employ adapter-based modules to inject conditional information, cross-attention control often suffers from inefficient semantic utilization and potential modality conflicts. To mitigate these issues, methods such as CreateLayout~\cite{zhang2025creatilayout} introduce siamese multimodal DiTs to decouple modalities and reduce interference. However, for RS imagery, the substantial domain gap between aerial and natural scenes, together with the absence of DiT models pre-trained on RS data, makes it highly uncertain to establish an effective mask-to-image control scheme.

In addition, the inherent stochasticity of the sampling process often leads to semantic drift and control deviation, resulting in unstable mask-conditioned generation and reduced effectiveness of synthetic data in downstream tasks. Recent studies typically adopt sample filtering strategies to directly select ~\cite{tang2025training}
or selectively utilize high-quality synthetic data~\cite{nguyen2023dataset}.
For instance, FreeMask~\cite{yang2023freemask} introduces adaptive filtering and re-sampling to suppress noisy labels caused by inconsistencies between generated images and masks, while promoting the synthesis of more complex scenes. Such post-processing strategies practically retain the effective regions of synthetic data for downstream training. Similarly, EarthSynth~\cite{pan2025earthsynth} employs a rule-based CLIP scoring mechanism~\cite{beaumont2022laion, gadre2023datacomp} to evaluate image–object–background triplets and filter more informative samples for downstream tasks.
Although these methods help reduce noise in synthetic datasets, they are inherently post-generation solutions and cannot fully resolve the intrinsic uncertainty of generation quality. In complex or few-shot scenarios, strict filtering may remove useful annotations and significantly degrade the performance of downstream tasks.

To address these limitations, we propose a task-oriented data synthesis framework that systematically examines mask-control strategies and introduces a control-rectify sampling method for RS semantic segmentation. We first compare different mask-to-image control schemes and observe that an appropriate DiT-based control design can markedly improve the effectiveness of synthetic data. In RS scenarios where fine-grained text descriptions are scarce, the text–image–mask joint attention scheme consistently outperforms both mask-adapter and siamese MM-attention approaches. 

Second, to mitigate the inherent uncertainty in sampling, we propose a control-rectify flow matching method, a plug-and-play sampling strategy guided by downstream task feedback. By leveraging the semantic loss signals from a pre-trained segmentation model, the sampler dynamically adjusts the sampling trajectory during the early high-plasticity stage, enabling in-process quality optimization. This improves the relevance and stability of synthetic data in complex and few-shot scenarios, rather than relying solely on post-processing filtering.

We conduct extensive experiments demonstrating that the proposed task-oriented data synthesis approach achieves OA/mIoU/mAcc improvements of $1.39\%$/$4.14\%$/$6.83\%$ on FUSU-4k~\cite{yuan2024fusu} and $1.60\%$/$2.08\%$/$2.22\%$ on LoveDA-5k~\cite{wang2021loveda} compared to baseline. Our main contributions are summarized as follows:

\begin{itemize}[label=\textbullet, leftmargin=2em]
\item We introduce TODSynth, a task-oriented data synthesis framework that integrates architecture-level and sampling-level control to enhance the effectiveness of synthetic data for RS semantic segmentation.

\item We propose a control-rectify flow matching (CRFM) method to mitigate generation uncertainty and enhance the performance in complex and few-shot RS scenes. It is a task-feedback–guided sampling optimization strategy that leverages a pre-trained task network to refine generation without retraining the diffusion model.

\item We conduct extensive experiments on two representative remote sensing semantic segmentation tasks, showing that the proposed TODSynth outperforms existing SOTA generation methods.

\end{itemize}

\section{Related Work}
\label{sec:formatting}
\subsection{Conditional Image Generation}

Conditional image generation~\cite{wang2025unicombine, zhang2025creatilayout, zhang2023adding, xie2024omnicontrol, mou2024t2i} aims to generate images under specific constraints such as layouts, masks, or text prompts. In this work, we mainly introduce layout‑to‑image and mask‑to‑image generation. ControlNet~\cite{zhang2023adding} achieves effective control by combining a frozen pre-trained diffusion backbone with a trainable copy through ``zero convolution" layers, enabling guidance of the decoder under single or multiple conditions, such as depth maps, edges, or segmentation masks. Beyond this, adapter-based variants such as UniControl~\cite{xie2024omnicontrol} enable flexible multi-condition fusion. Recent MM-DiTs~\cite{esser2024scaling, wu2025less, tan2025ominicontrol} explore different control strategies based on DiT architecture.
For example, CreateLayout~\cite{zhang2025creatilayout} introduces a siamese MM-DiT to decouple text and layout modalities, while EasyControl~\cite{zhang2025easycontrol} and UniCombine~\cite{wang2025unicombine} employ lightweight condition injection and use LoRA modules to improve training efficiency.

Synthesizing  training data for remote sensing (RS) semantic segmentation~\cite{DONG2026103839, nguyen2023dataset, chen2019learning} has gained increasing attention due to the scarcity of labeled data. SatSynth~\cite{toker2024satsynth} explores joint generation of images and corresponding masks for satellite segmentation by training diffusion models from scratch. However, training diffusion models on small RS datasets often yields suboptimal results. To overcome this limitation, several RS generative foundation models have been developed, including DiffusionSat~\cite{khanna2023diffusionsat}, EarthSynth~\cite{pan2025earthsynth}, and TerraGen~\cite{tang2025terragen}, which build upon Stable Diffusion (SD) v1.5~\cite{stable-diffusion-v1-5} and are trained on large RS datasets to enable controllable generation. Similarly, TISynth~\cite{DONG2026103839} introduces additional reference guidance to achieve transferable RS semantic segmentation data synthesis via joint reference-semantic fusion, also based on SD v1.5.

Despite these advances, major challenges remain due to the domain gap between natural and RS imagery, the limited availability of RS-specific DiT pre-trained models, and the uncertainty in effective DiT control strategies. In this work, we systematically explore mask-to-image control schemes for RS semantic segmentation and adopt MM-DiT with unified triple attention.

\subsection{Training Data Synthesis for Downstream Tasks}

Synthetic data has been widely employed to augment training datasets for downstream tasks such as semantic segmentation. Existing works~\cite{nguyen2023dataset} generally adopt uncertainty-aware segmentation approaches
or sample filtering methods, e.g., adaptive filtering~\cite{yang2023freemask} and CLIP-based selection~\cite{pan2025earthsynth, tang2025training}.
While these methods are effective in reducing noise, they rely on post-processing and cannot resolve the intrinsic uncertainty of generation quality, particularly in complex or few-shot scenarios.

In related areas such as dataset distillation~\cite{wang2025dataset,sajedi2023datadam,cui2023scaling} and layout-to-image (L2I) generation~\cite{zheng2023layoutdiffusion,zhang2025creatilayout}, some works attempt to improve synthetic data via distribution matching~\cite{kim2022dataset, lee2022dataset} or trajectory matching~\cite{du2023minimizing,cazenavette2022dataset}. The training-free L2I method~\cite{chen2025trainingfreelayouttoimagegenerationmarginal} apply marginal attention constraints to optimize latent features, enhancing spatial controllability and reducing semantic failures.
However, these strategies are misaligned with semantic segmentation objectives, limiting their direct applicability.

Different from these approaches, we propose a control-rectify flow matching method to mitigate generation uncertainty and enhance the effectiveness of synthetic data for downstream tasks. Our method leverages semantic loss to dynamically adjust the sampling directions during the early, high-plasticity stage of generation. This strategy improves the effectiveness of generated data without directly modifying feature maps or constraining the distribution of synthetic samples, providing a task-oriented solution for robust data synthesis in RS semantic segmentation. Additional related work on inversion-free editing is provided in Appendix~\ref{sec:discussions}.

\section{Methodology}

\begin{figure*}[t]
 \centering
 \includegraphics[width=1.0\linewidth]{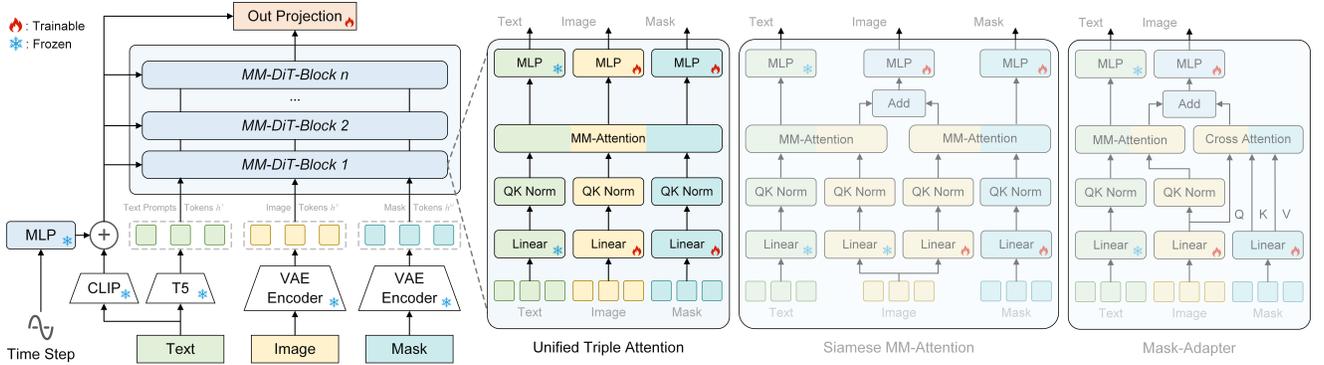}
 \caption{The architecture of MM-DiT with unified triple attention. For comparison, the mask-adapter and siamese MM-attention schemes are also illustrated.}
 \label{fig:2}
 \vspace{-4mm}
\end{figure*}

\subsection{Preliminary}

Flow-based generative models formulate data synthesis as a continuous transport from a base distribution to the target distribution. The evolution of a sample along a trajectory $\mathbf{y}_t$ parameterized by $t\in[0,1]$ is described by an ordinary differential equation (ODE):
\begin{equation}
\frac{d\mathbf{y}_t}{dt} = \mathbf{v}_\theta(\mathbf{y}_t, t),
\end{equation}
where $\mathbf{v}_\theta$ denotes a learnable velocity field. Flow Matching (FM)~\cite{lipman2022flow} supervises the model with ideal transport velocities constructed between paired samples, enabling stable training without stochastic score matching. Rectified Flow (RF)~\cite{liu2022flow} further simplifies this by assuming a linear trajectory between the endpoints:
\begin{equation}
\mathbf{z}_t = (1-t)\mathbf{z}_0 + t\mathbf{z}_1,
\end{equation}
with constant velocity $\mathbf{v}_\theta = \mathbf{z}_1 - \mathbf{z}_0$, allowing deterministic and efficient generation.



\subsection{The Overall Workflow of TODSynth}

Due to the high cost of expanding RS datasets, we propose TODSynth, a training data synthesis framework designed to enhance the performance of downstream RS semantic segmentation models. Considering the challenges of fine-grained semantic control, we integrate architecture-level and sampling-level control strategies to improve the effectiveness of synthetic data. The overall workflow is illustrated in Fig.~\ref{fig:1}.

At the architecture level, to enhance mask-to-image control, we adopt the MM-DiT architecture combined with the pre-trained SD v3.5 model (Fig.~\ref{fig:1}(a)). To address semantic imbalance, global semantic information is embedded into the text prompt, which is combined with pixel-level mask guidance. A text–image–mask unified attention mechanism is employed to strengthen the model’s ability to generate complex scenes and few-shot categories. We further compare different DiT-based control strategies (Fig.~\ref{fig:2}), demonstrating that the text–image–mask joint attention scheme consistently outperforms both mask-adapter and siamese MM-attention approaches on this task.

Despite improvements at the architecture level, the stochastic nature of the generation trajectory can still cause deviations from the semantic mask constraints in certain regions, reducing the effectiveness of the synthesized data. To address this, we propose a sampling-level control strategy, namely Control-Rectify Flow Matching (CRFM) (Fig.~\ref{fig:1}(b)). During the high-plasticity early sampling stage, CRFM adjusts the trajectory by applying a rectification vector $\mathbf{v}_{\text{rec}}'$ to the predicted velocity field. This vector is computed from the cross-entropy loss between early predictions and the ground-truth mask, enhancing the alignment of the generated samples with the semantic constraints and mitigating uncertainty in data synthesis.

Finally, the refined synthetic data is combined with real samples and used to train downstream semantic segmentation models (Fig.~\ref{fig:1}(c)).


\subsection{Unified Triple Attention}

Our work builds upon the SOTA MM-DiT architecture adopted by the SD v3.5 family. We focus on effectively incorporating semantic masks $\mathbf{C}^m$ to achieve precise spatial control during image generation. MM-DiT performs multimodal fusion using a Unified Attention mechanism that processes the text sequence $\mathbf{h}^t$ and image latent sequence $\mathbf{h}^z$ jointly by concatenating their tokens:
\begin{equation}
\mathbf{h^t_o},\mathbf{h^z_o} = Attn([\mathbf{h^t}, \mathbf{h^z}]),
\end{equation}
where $[.,.]$ denotes length-wise concatenation. Each modality uses its own projection matrices ($\mathbf{W}_q$, $\mathbf{W}_k$, $\mathbf{W}_v$) to map features into the query, key, and value spaces. On this basis, we examine three mask-conditioning strategies: unified triple attention, siamese MM-attention, and mask-adapter.


Unified triple attention (Tri-attention) extends MM-DiT by adding a third modality stream for the mask sequence $\mathbf{h}^m$. Following the unified fusion principle, attention is computed over all three token sequences:
\begin{equation}
\begin{aligned}
\mathbf{h^t_o}, \mathbf{h^z_o}, \mathbf{h^m_o}= Attn([\mathbf{h^t}W^t_q, \mathbf{h^z}W^z_q, \mathbf{h^m}W^m_q],\\
[\mathbf{h^t}W^t_k, \mathbf{h^z}W^z_k, \mathbf{h^m}W^m_k],\\
[\mathbf{h^t}W^t_v, \mathbf{h^z}W^z_v, \mathbf{h^m}W^m_v]).
\end{aligned}
\end{equation}
Tri-attention integrates the mask stream directly, enabling crucial cross-modal interaction between the textual and mask modalities during the fusion process, allowing the semantic information encoded in the text to enhance the model's understanding of the required spatial and semantic distribution.

We also evaluate the siamese MM-attention in L2I~\cite{zhang2025creatilayout}. It processes text–image and mask–image interactions through two separate MM-attention blocks.
Unlike L2I generation, the pure segmentation mask in M2I lacks localized textual descriptions, reducing semantic richness. Although the siamese design originally intended to mitigate token-length imbalance across modalities, this issue is less pronounced in M2I due to the natural tokenization of the mask, diminishing the benefit of decoupled attention.

Mask-Adapter injects mask features into the image branch using a lightweight cross-attention module while keeping the MM-DiT backbone unchanged.
Although computationally efficient, the mask information suffers from two limitations: 1) it is not directly fused with text embeddings, weakening the learning of global semantic information, especially in fine-grained and few-shot scenarios. 2) The mechanism only allows $\mathbf{h}^z$ to query $\mathbf{h}^m$, meaning the mask representation remains static throughout the denoising process, restricting its adaptive refinement. These constraints limit the precision and controllability of the generated results.

Through the above analysis, we adopt the Tri-attention scheme and fully fine-tune both the image and mask branches, given the lack of DiT-based generative models pre-trained on diverse RS imagery.

\subsection{Control-Rectify Flow Matching}

Upon training a conditional generative flow model~\cite{lipman2022flow,liu2022flow}, the inherent stochasticity of the trajectory and the imperfections in model prediction inevitably lead to generated images that, in certain regions, deviate from the semantic mask constraints during the inference process. Such conditional deviations often manifest in the early stages of the model's generation, where small errors are rapidly amplified. Therefore, we propose the Control Rectify Flow Matching (CRFM) mechanism, enabling the timely correction of the velocity vector early in the flow trajectory, thereby ensuring the synthesized results adhere more faithfully to the target semantic distribution.

The original RF model predicts a constant vector field, defined simply as $\mathbf{v} = \mathbf{z}_1 - \mathbf{z}_0$. When incorporating the conditional controls $\mathbf{C}^t$ and $\mathbf{C}^m$ (text and semantic mask), the ideal vector field $\mathbf{v}_\Theta$ for the conditional flow trajectory retains this straight-line property. It can be formally expressed as:
\begin{equation}
\mathbf{v}_\Theta(\mathbf{z}_t, t, \mathbf{C}^t, \mathbf{C}^m) = \mathbf{z}_1 - \mathbf{z}_0.
\end{equation}
This formulation implies that, ideally, at every time step $t$ along the probabilistic path, the model's predicted vector field remains the deterministic constant difference between the terminal data point $\mathbf{z}_0$ and the initial noise point $\mathbf{z}_1$, irrespective of the current state $\mathbf{z}_t$.

Therefore, the generation process, which maps the initial noise sample $\mathbf{z}_1$ to the final data sample $\mathbf{z}_0$, is described by solving the ODE via integration from $t=1$ to $t=0$:
\begin{equation}
\mathbf{z}_0 = \mathbf{z}_1 + \int_{1}^{0} \mathbf{v}_\Theta(\mathbf{z}_t, t, \mathbf{C}^t, \mathbf{C}^m) \, dt.
\end{equation}
In practical implementation, this continuous integration is typically approximated using numerical solvers, such as the Euler method, which discretizes the process into $N$ time steps. The iterative generation is then described by the following update rule:
\begin{equation}
\mathbf{z}_{t_{i-1}} = \mathbf{z}_{t_i} + \mathbf{v}_\Theta(\mathbf{z}_{t_i}, t_i, \mathbf{C}^t, \mathbf{C}^m) \cdot \Delta t,
\end{equation}
where $\mathbf{z}_{t_N} = \mathbf{z}_1$ is the starting noise, and the final generated sample is $\mathbf{z}_{t_0} \approx \mathbf{z}_0$.


Assuming $\mathbf{v}_\Theta^T$ is the target velocity field that guides samples toward the data distribution, and $\mathbf{v}_\Theta^P$ is the learned velocity field predicted by the flow model, the residual vector $\mathbf{v}_{\text{res}}$ can be calculated as the difference between the true and predicted fields:
\begin{equation}
\mathbf{v}_{\text{res}} = \mathbf{v}_\Theta^T - \mathbf{v}_\Theta^P.
\end{equation}
We utilize this residual vector as the rectified vector field $\mathbf{v}_{\text{rec}}$. Consequently, the target velocity field can be decomposed and represented as:
\begin{equation}
\mathbf{v}_\Theta = \mathbf{v}_\Theta^P + \mathbf{v}_{\text{rec}}.
\end{equation}
Theoretically, if we can accurately compute $\mathbf{v}_{\text{rec}}$ at every step along the flow trajectory, then regardless of any inherent drift or deviation in the learned field, the final synthesized result will align well with the target data distribution.

Unlike standard data distributions, the conditional data distribution $q(\mathbf{z}_0|(\mathbf{C}^t, \mathbf{C}^m))$ represents the mapping from the initial latent noise $\mathbf{z}_1 \sim p(\mathbf{z})$ to the target latent state $\mathbf{z}_0$, conditioned on textual prompts $\mathbf{C}^t$ and semantic masks $\mathbf{C}^m$. Generating samples from this conditional distribution requires the flow model to produce a velocity field that strictly respects the specified conditions, particularly the spatial constraints imposed by $\mathbf{C}^m$. Although the unified triple attention scheme is designed to produce an appropriate conditional vector field based on the input modalities, the stochasticity inherent in the initial latent noise and the complexity of the mask constraints inevitably lead to deviations in the generated semantics. This motivates the need for an early, targeted trajectory correction mechanism to enforce strict adherence to the conditions throughout the flow.

The proposed CRFM mechanism introduces an innovative, targeted gradient-based correction at the early stages of the generation trajectory. Specifically, CRFM first generates a pre-synthesized prediction of the final sample, $\mathbf{x}_0^t$, using the current state, $\mathbf{z}_0$ and the predicted velocity field $\mathbf{v^P}$:
\begin{equation}
\mathbf{z}_0^t = \mathbf{z}_t - \sigma_t \mathbf{v^P}(\mathbf{z}_t, t, \mathbf{C}^t, \mathbf{C}^m),
\end{equation}
\begin{equation}
\mathbf{x}_0^t = \mathcal{D}(\mathbf{z}_0^t),
\end{equation}
where $\mathcal{D}$ denotes the VAE decoder that maps the latent space to the image space. Here, $\mathbf{z}_0^t$ is the predicted latent representation of the final sample, and $\mathbf{x}_0^t$ serves as the input to a pre-trained segmentation network $\mathcal{S}$. The output of this network is then used to compute a Cross-Entropy loss against the ground-truth semantic mask $\mathbf{C}^m$:
\begin{equation}
\mathbf{v}_{\text{rec}}' = -\nabla_{v_t} \mathcal{L}_{CE}(\mathcal{S}(\mathbf{x}_0^t), \mathbf{C}^m).
\end{equation}
The resulting gradient is back-propagated to produce the predicted rectified vector field, denoted as $\mathbf{v}_{\text{rec}}'$, which estimates the theoretically required residual vector $\mathbf{v}_{\text{rec}}$. This correction is incorporated to update the predicted velocity field:
\begin{equation}
\mathbf{v}'(\mathbf{z}_t, t, \mathbf{C}^t, \mathbf{C}^m) = \mathbf{v^P}(\mathbf{z}_t, t,
\mathbf{C}^t, \mathbf{C}^m) + \alpha \mathbf{v}_{\text{rec}}',
\end{equation}
where $\alpha$ is a tunable scalar that controls the strength of the correction applied at each step.
This process effectively adjusts the predicted velocity vector $\mathbf{v}^P$ during the critical early steps, ensuring stricter adherence to the specified conditions throughout the generation. The proposed CRFM differs fundamentally from previous methods in its optimization target. Instead of performing gradient optimization directly on the latent state $\mathbf{z}_t$~\cite{chen2025trainingfreelayouttoimagegenerationmarginal}, we focus on rectifying the underlying velocity vector field $\mathbf{v}_\Theta$, which indirectly updates $\mathbf{z}_t$ through ODE integration. 

Based on the above theoretical analysis, we also validated this intuition experimentally. We found that direct optimization of $\mathbf{z}_t$ using the gradient of downstream tasks leads to mode collapse, where the generative model fails to capture the full diversity of the target distribution and instead converges to a limited subset of samples, resulting in ineffective synthetic data. By contrast, optimizing the underlying velocity field provides stable, continuous corrections along the trajectory, thereby effectively mitigating mode collapse.

Moreover, it is important to note that we apply this task-guided loss gradient only during the early stages of sampling. On one hand, the early stage exhibits high stochasticity, making it more amenable to trajectory adjustment. On the other hand, pre-trained segmentation models inherently contain prediction errors. These errors, although present, can still provide useful guidance for coarse adjustments at early steps. Applying corrections too late in the trajectory, however, could amplify these errors, making the sampling process overly dependent on the segmentation model. This may generate adversarial-like perturbations that deviate from the normal image distribution and degrade synthetic quality. Therefore, the proposed CRFM adjusts the directions of velocity field only in the early step, effectively reducing uncertainty in data synthesis and ensuring that the generated results adhere more faithfully to the target semantic distribution.

\subsection{Implementation Details}

Our TODSynth model leverages the MM-DiT~\cite{esser2024scaling} architecture, building upon the foundational SD v3.5~\cite{stabilityai_sd35_release_2024} framework. The model is trained at a $512 \times 512$ resolution. Optimization is performed using the AdamW optimizer, setting the learning rate to $\eta = 1 \times 10^{-5}$ and the weight decay to $0.01$. For post-processing, image synthesis for each semantic mask involves redefining and fixing a sequence of random seeds. A subsequent filtering step ensured that the synthesized images either resulted in a minimum of three distinct classes or preserved those classes containing only a few samples. Our models are trained for $200,000$ steps on 8 NVIDIA RTX 4090 GPUs.

Our semantic segmentation approach is implemented using the MMSegmentation framework~\cite{mmseg2020}. For fair comparison and robust training, we adopt the standard data augmentation techniques integrated within MMSegmentation. These techniques include basic transformations (random crop and random flip) and the comprehensive PhotoMetricDistortion, which sequentially applies a range of photometric adjustments, such as random brightness, random contrast, random saturation, and random hue.

\label{sec:formatting}

\begin{table*}
\centering
\caption{Quantitative comparison with different training data synthesis methods (\%). We report segmentation performance, the use of post-processing filters, and the ratio of synthetic to real training data for each method.}
\label{tab:1}
\begin{tabular}{lccccccccc}
\toprule
{Method} & {Post-processing} & {Syn/Real volume} & \multicolumn{3}{c}{FUSU-4k} & \multicolumn{3}{c}{LOVEDA-5k} \\
\cmidrule(lr){4-6} \cmidrule(lr){7-9}
& & & OA & mIoU & mAcc & OA & mIoU & mAcc \\
\midrule
Baseline~\cite{zhao2017pyramid} & - & - & 74.27 & 45.27 & 56.44 & 78.22 & 64.04 & 78.71 \\
Freestyle~\cite{xue2023freestyle} & $\times$ & $\times$10 & 73.49 & 44.99 & 57.75 & 78.59 & 64.10 & 78.77 \\
Controlnet~\cite{zhao2023label} (SD v1.5) & $\times$ & $\times$10 & 73.85 & 45.13 & 56.77 & 78.24 & 63.89 & 79.42 \\
FreeMask~\cite{yang2023freemask} & FM & $\times$5 & 74.23 & 45.83 & 56.29 & 79.26 & 64.93 & 79.06 \\
SynthEarth*~\cite{pan2025earthsynth} & CLIP Score & $\times$5 & 75.35 & 47.53 & 58.91 & 78.87 & 64.07 & 78.00 \\
\midrule
SD v3.5~\cite{stabilityai_sd35_release_2024}& FM & $\times$3 & 75.41 & 48.57 & 61.67 & 79.72 & 65.45 & 79.91 \\
Ours & FM & $\times$3 & \textbf{75.66} & \textbf{49.41} & \textbf{63.27} & \textbf{79.82} & \textbf{66.12} & \textbf{80.93} \\
\bottomrule
\multicolumn{9}{l}{\footnotesize $^*$SynthEarth is a unified generative model that can be directly used for inference.}
\end{tabular}
\vspace{-4mm}
\end{table*}

\begin{table}
\centering
\caption{Comparison of different control strategies on FUSU-4k.}
\label{tab:fusu_comparison}
\begin{tabular}{lccc}
\toprule
{Method} & \multicolumn{3}{c}{FUSU-4k} \\
\cmidrule(lr){2-4}
& OA & mIoU & mAcc \\
\midrule
Controlnet~\cite{zhao2023label} (SD v1.5) & 73.85 & 45.13 & 56.77 \\
Mask-adapter & 74.94 & 47.41 & 59.62 \\
Siamese MM-attention & 74.94 & 48.46 & 61.44 \\
Tri-Attention & \textbf{75.41} & \textbf{48.57} & \textbf{61.67} \\
\bottomrule
\end{tabular}
\vspace{-4mm}
\end{table}

\begin{table*}[t]
\centering
\caption{Comparison of different post-processing strategies on FUSU-4k.}
\label{tab:3}
\begin{tabular}{c c c c c c c}
\toprule
\multicolumn{1}{c}{Experiment} & 
\multicolumn{1}{c}{FreeMask Filter~\cite{yang2023freemask}} & 
\multicolumn{1}{c}{CLIP Score Filter~\cite{pan2025earthsynth}} & 
\multicolumn{1}{c}{CRFM} & 
\multicolumn{1}{c}{OA} & 
\multicolumn{1}{c}{mIoU} & 
\multicolumn{1}{c}{mAcc} \\
\midrule
Exp1 & $\times$ & $\checkmark$ & $\times$ & 74.78 & 46.69 & 58.08 \\
Exp2 & $\checkmark$ & $\times$ & $\times$ & 75.41 & 48.57 & 61.67 \\
Exp3 (Ours) & $\checkmark$ & $\times$ & $\checkmark$ & \textbf{75.66} & \textbf{49.41} & \textbf{63.27} \\
\bottomrule
\end{tabular}
\vspace{-4mm}
\end{table*}

\section{Experiment}

\subsection{Datasets and Evaluation}

\noindent\textbf{FUSU}. FUSU~\cite{yuan2024fusu} is a high–resolution aerial land-cover dataset originally developed for change detection research. The imagery is sourced from Google Earth with a spatial resolution of approximately 0.2–0.5 m. FUSU provides fine-grained annotations covering 17 land-cover categories (e.g., residential, industrial, cropland, inland water, wetland, park, commercial land, greenbelt, service land, etc.). Following the setting in TISynth~\cite{DONG2026103839}, we use a 4,000-sample subset of FUSU (denoted as FUSU-4k) and adopt the T1 image–mask pairs cropped to $512 \times 512$ pixels for training and evaluation.

\noindent\textbf{LoveDA}. LoveDA~\cite{wang2021loveda} is a domain-adaptive semantic segmentation benchmark consisting of 5,987 images with 0.3 m resolution and a spatial size of $1,024\times1,024$ pixels. It includes six semantic classes: building, road, water, barren, forest, and agriculture. In this work, we consider the rural-to-urban setting, using 3,274 rural images for training and 2,713 urban images for testing.

\noindent\textbf{Evaluation metrics}. Semantic segmentation performance is measured using Overall Accuracy (OA), Mean Intersection over Union (mIoU), and Mean Accuracy (mAcc). Image synthesis quality is evaluated using Frechet Inception Distance (FID).

\subsection{Comparison Results}

We compare TODSynth with SOTA training data synthesis methods on two RS semantic segmentation datasets. The comparison methods include four UNet-based controllable generators(i.e., Freestyle~\cite{xue2023freestyle}, FreeMask~\cite{yang2023freemask}, ControlNet~\cite{zhang2023adding}, and SynthEarth~\cite{pan2025earthsynth}) and a MM-DiT-based controllable generation method (i.e., SD v3.5 with Tri-attention).

For a fair comparison, all methods except SynthEarth are fine-tuned on the same RS datasets. SynthEarth, as a large-scale generative foundation model trained on extensive RS imagery and supporting training-free controllable generation, is directly used to synthesize RS segmentation data without additional tuning. All downstream semantic segmentation experiments use the same model architecture.

To isolate the contribution of the proposed CRFM, we evaluate SD v3.5 using the same architecture and post-processing filter as TODSynth. Therefore, the presence or absence of CRFM is the only difference. Table~\ref{tab:1} summarizes distinctions among among all methods, including post-processing strategies, the ratio of synthetic to real training samples, and segmentation performance on FUSU-4k and LoveDA. Fully supervised training on real data is used as the baseline.

Across both datasets, TODSynth consistently achieves the best performance. Compared to the supervised baseline trained on real data, TODSynth improves OA/mIoU/mAcc by $1.39\%$/$4.14\%$/$6.83\%$ on FUSU-4k and $1.60\%$/$2.08\%$/$2.22\%$ on LoveDA. Moreover, MM-DiT–based models exhibit stronger controllability compared with UNet-based generators, even when the SynthEarth is a RS generative foundation model. This advantage is particularly evident on FUSU, which contains 17 diverse land-cover categories.

Comparing TODSynth with SD v3.5 under identical architecture and post-processing settings, our CRFM achieves an improvement of $0.84\%$/$1.6\%$ mIoU/mAcc on FUSU-4k and $0.67\%$/$1.02\%$ mIoU/mAcc on LoveDA by dynamically adjusting the sampling trajectory. This demonstrates the effectiveness of the proposed sampling-level control. 

Furthermore, TODSynth achieve superior performance with fewer synthetic samples compared to methods requiring larger synthetic datasets, highlighting its practical utility.

\subsection{Ablation Study}

We conduct the ablation study on the FUSU dataset. All experiments use the same post-processing strategy and downstream semantic segmentation implementation.

For DiT-based controllable generation, we compare three control strategies: Unified Triple Attention (Tri-Attention), Siamese MM-Attention, and Mask-Adapter, using ControlNet (SD v1.5) as the baseline. As shown in Table 2, Tri-Attention consistently outperforms both Siamese MM-Attention and Mask-Adapter. The Mask-Adapter, while efficient, provides limited control because it does not fuse text and mask embeddings and keeps mask features static, restricting precision and adaptability. In contrast, MM-DiT–based methods achieve better performance, and Tri-Attention achieves the best results with a concise implementation. We therefore adopt Tri-Attention in this work.

For post-processing of synthetic data, current approaches can be grouped into image-level filtering, e.g., CLIP-based methods like Earthsynth~\cite{pan2025earthsynth} and SDS~\cite{tang2025training}, and pixel-level filtering, e.g., Freemask~\cite{yang2023freemask} and TISynth~\cite{DONG2026103839}. We compare these two types of post-processing filters, as well as the combination with our proposed CRFM. As shown in Table~\ref{tab:3}, pixel-level filtering significantly improves downstream performance, particularly in mIoU and mAcc, due to its finer-grained region selection. Incorporating CRFM into the pixel-level filter further boosts performance ($0.84\%$/$1.60\%$ mIoU/mAcc on FUSU), demonstrating its effectiveness in complex and few-shot scenarios.

\begin{table}
\centering
\caption{Performance of CRFM with different sampling steps and CRFM steps on FUSU-4k.}
\label{tab:4}
\begin{tabular}{lccccc}
\toprule
Method & OA & mIoU & mAcc & FID \\
\midrule
step=23, CRFM=0 & 75.41 & 48.57 & 61.67 & 35.85 \\
step=23, CRFM=2 & \textbf{75.79} & 48.80 & 61.05 & \textbf{34.86} \\
step=23, CRFM=4 & 75.66 & \textbf{49.41} & \textbf{63.27} & 38.65 \\
step=23, CRFM=6 & 75.71 & 48.74 & 61.30 & 66.95  \\
\midrule
step=18, CRFM=0 & 75.53 & 47.81 & 59.32 & 40.23  \\
step=18, CRFM=2 & 75.86 & 47.99 & 59.46 & \textbf{38.94}  \\
step=18, CRFM=4 & \textbf{76.24} & 48.93 & 60.54 & 57.01  \\
step=18, CRFM=6 & 76.02 & \textbf{49.57} & \textbf{62.19} & 86.69  \\
\midrule
step=13, CRFM=0 & 75.02 & 46.86 & 59.44 & 49.53  \\
step=13, CRFM=2 & 75.73 & 48.29 & 59.99 & \textbf{48.61}  \\
step=13, CRFM=4 & 76.08 & 49.04 & 62.05 & 81.90  \\
step=13, CRFM=6 & \textbf{76.19} & \textbf{49.09} & \textbf{62.50} & 132.92  \\
\bottomrule
\end{tabular}
\end{table}

\subsection{Discussion}

We further analyze the effectiveness and sensitivity of CRFM. We conduct extensive experiments using different sampling steps and CRFM steps to test its sensitivity to the sampling step and empirical usage.

As discussed in the Methodology section, early adjustment of the sampling trajectory helps reduce the gap between synthetic data and downstream tasks, mitigating the uncertainty in synthetic data. First, with the standard sampling step of 23, we apply CRFM starting from the first step, using 2, 4, and 6 CRFM steps. CRFM=0 represents the baseline without CRFM. As shown in Table~\ref{tab:4}, we observe that the effectiveness of synthetic data for downstream tasks and its quality initially improve and then decline as CRFM steps increase. This is because early-stage randomness allows easier trajectory adjustment, but excessive adjustment becomes constrained by the capabilities of the pre-trained segmentation model. Over-adjustment also generates adversarial-like samples that deviate from the natural image distribution, causing a deterioration in FID. Therefore, CRFM is sensitive to the number of adjustment steps and the pre-trained segmentation model. In practice, applying CRFM for about the first four steps balances downstream performance and synthetic data quality.

Next, we test shorter sampling steps with 18 and 13 steps, again starting CRFM from the first step with 2, 4, and 6 adjustment steps. We observe the same trend: synthetic data quality initially improves with more CRFM steps but declines if over-adjusted, particularly at 6 steps, where downstream performance may still increase but FID drops significantly due to adversarial-like artifacts. Thus, when reducing the number of sampling steps, we also reduce the number of CRFM steps. Additional sensitivity analyses regarding the pre-trained segmentation model appear in the Appendix~\ref{sec:experiments}.

Based on these analyses, we summarize the limitations and future directions of this study:

The first issue is CRFM sensitivity. CRFM is sensitive to the sampling step and the pre-trained segmentation model. Applying it for a small number of early steps is the most stable approach. It also requires a sufficiently capable pre-trained segmentation model; with very limited data, the model does not provide effective guidance, reducing CRFM’s effectiveness. We will explore more robust pre-trained models, such as RS foundation models, to improve the effective of evaluation and sampling stability.

The second issue is synthetic data diversity and cross-domain performance. Currently, we do not leverage a strong RS generative foundation model or external RS reference data, which limits the diversity of synthetic samples. Future work explores incorporating additional information to enhance the effectiveness of synthetic data for downstream tasks.

\section{Conclusion}

In this paper, we propose a training data synthesis framework for RS semantic segmentation that integrates an MM-DiT architecture with task-guided sampling. We systematically evaluate different mask-to-image control schemes for this task. To enhance control over mask boundaries and semantics in the synthesized data, we adopt a unified triple attention mechanism within MM-DiT control blocks, improving the interaction between text, image, and mask. Furthermore, we introduce the control-rectify flow matching method, guided by semantic loss, to dynamically adjust sampling directions during the early, high-plasticity stage. This approach mitigates the intrinsic uncertainty of generation quality, thereby improving the utility of synthesized data for downstream semantic segmentation tasks. Extensive experiments demonstrate the strong control capabilities of our framework in mask-guided synthesis and task-guided sampling, surpassing SOTA approaches for semantic segmentation data synthesis. This work shows that task-oriented data synthesis can substantially enhance the effectiveness of synthetic data for downstream applications.

\section*{Acknowledgement}
This research was supported in part by the National Natural Science Foundation of China (Grant No. T2125006 and No. 42301390)

{
    \small
    \bibliographystyle{ieeenat_fullname}
    \bibliography{main}

@String(AAAI = {AAAI})

@inproceedings{duan2025dit4sr,
  title={Dit4sr: Taming diffusion transformer for real-world image super-resolution},
  author={Duan, Zheng-Peng and Zhang, Jiawei and Jin, Xin and Zhang, Ziheng and Xiong, Zheng and Zou, Dongqing and Ren, Jimmy S and Guo, Chunle and Li, Chongyi},
  booktitle={Proceedings of the IEEE/CVF International Conference on Computer Vision},
  pages={18948--18958},
  year={2025}
}

@inproceedings{tang2025training,
  title={A Training-free Synthetic Data Selection Method for Semantic Segmentation},
  author={Tang, Hao and Yu, Siyue and Pang, Jian and Zhang, Bingfeng},
  booktitle={Proceedings of the AAAI Conference on Artificial Intelligence},
  volume={39},
  number={7},
  pages={7229--7237},
  year={2025}
}

@article{nguyen2023dataset,
  title={Dataset diffusion: Diffusion-based synthetic data generation for pixel-level semantic segmentation},
  author={Nguyen, Quang and Vu, Truong and Tran, Anh and Nguyen, Khoi},
  journal={Advances in Neural Information Processing Systems},
  volume={36},
  pages={76872--76892},
  year={2023}
}

@article{pan2025earthsynth,
  title={EarthSynth: Generating Informative Earth Observation with Diffusion Models},
  author={Pan, Jiancheng and Lei, Shiye and Fu, Yuqian and Li, Jiahao and Liu, Yanxing and Sun, Yuze and He, Xiao and Peng, Long and Huang, Xiaomeng and Zhao, Bo},
  journal={arXiv preprint arXiv:2505.12108},
  year={2025}
}

@inproceedings{xie2024omnicontrol,
    title={OmniControl: Control Any Joint at Any Time for Human Motion Generation},
    author={Yiming Xie and Varun Jampani and Lei Zhong and Deqing Sun and Huaizu Jiang},
    booktitle={The Twelfth International Conference on Learning Representations},
    year={2024},
    url={https://openreview.net/forum?id=gd0lAEtWso}
}

@misc{chen2025trainingfreelayouttoimagegenerationmarginal,
      title={Training-Free Layout-to-Image Generation with Marginal Attention Constraints}, 
      author={Huancheng Chen and Jingtao Li and Weiming Zhuang and Haris Vikalo and Lingjuan Lyu},
      year={2025},
      eprint={2411.10495},
      archivePrefix={arXiv},
      primaryClass={cs.CV},
      url={https://arxiv.org/abs/2411.10495}, 
}

@article{yang2023freemask,
  title={Freemask: Synthetic images with dense annotations make stronger segmentation models},
  author={Yang, Lihe and Xu, Xiaogang and Kang, Bingyi and Shi, Yinghuan and Zhao, Hengshuang},
  journal={Advances in Neural Information Processing Systems},
  volume={36},
  pages={18659--18675},
  year={2023}
}

@article{chen2022contrasting,
  title={Contrasting inequality in human exposure to greenspace between cities of Global North and Global South},
  author={Chen, Bin and Wu, Shengbiao and Song, Yimeng and Webster, Chris and Xu, Bing and Gong, Peng},
  journal={Nature Communications},
  volume={13},
  number={1},
  pages={4636},
  year={2022},
  publisher={Nature Publishing Group UK London}
}

@ARTICLE{11134495,
  author={Li, Ziming and Wang, Han and Wang, Yadian and Chen, Gang and Chen, Jing M. and Chen, Bin},
  journal={IEEE Transactions on Geoscience and Remote Sensing}, 
  title={Large-Scale High-Resolution Essential Urban Land Cover Category Mapping Using a Semantic-Augmented and Noise-Tolerant Approach}, 
  year={2025},
  volume={63},
  number={},
  pages={1-21},
  doi={10.1109/TGRS.2025.3601626}}

@inproceedings{li2025segearth,
  title={Segearth-ov: Towards training-free open-vocabulary segmentation for remote sensing images},
  author={Li, Kaiyu and Liu, Ruixun and Cao, Xiangyong and Bai, Xueru and Zhou, Feng and Meng, Deyu and Wang, Zhi},
  booktitle={Proceedings of the Computer Vision and Pattern Recognition Conference},
  pages={10545--10556},
  year={2025}
}

@inproceedings{dong2023large,
  title={Large-scale land cover mapping with fine-grained classes via class-aware semi-supervised semantic segmentation},
  author={Dong, Runmin and Mou, Lichao and Chen, Mengxuan and Li, Weijia and Tong, Xin-Yi and Yuan, Shuai and Zhang, Lixian and Zheng, Juepeng and Zhu, Xiaoxiang and Fu, Haohuan},
  booktitle={Proceedings of the IEEE/CVF International Conference on Computer Vision},
  pages={16783--16793},
  year={2023}
}

@article{DONG2026103839,
title = {Transferable image synthesis for remote sensing semantic segmentation via joint reference-semantic fusion},
journal = {Information Fusion},
volume = {127},
pages = {103839},
year = {2026},
issn = {1566-2535},
doi = {https://doi.org/10.1016/j.inffus.2025.103839},
url = {https://www.sciencedirect.com/science/article/pii/S1566253525009017},
author = {Runmin Dong and Shuai Yuan and Litong Feng and Jinxiao Zhang and Weijia Li and Mengxuan Chen and Bin Luo and Wayne Zhang and Haohuan Fu},
}

@inproceedings{namekata2024emerdiff,
	title={EmerDiff: Emerging Pixel-level Semantic Knowledge in Diffusion Models}, 
	author={Koichi Namekata and Amirmojtaba Sabour and Sanja Fidler and Seung Wook Kim},
	booktitle={The Twelfth International Conference on Learning Representations},
	year={2024},
  url={https://openreview.net/forum?id=YqyTXmF8Y2}
}

@misc{gao2024luminat2xtransformingtextmodality,
      title={Lumina-T2X: Transforming Text into Any Modality, Resolution, and Duration via Flow-based Large Diffusion Transformers}, 
      author={Peng Gao and Le Zhuo and Dongyang Liu and Ruoyi Du and Xu Luo and Longtian Qiu and Yuhang Zhang and Chen Lin and Rongjie Huang and Shijie Geng and Renrui Zhang and Junlin Xi and Wenqi Shao and Zhengkai Jiang and Tianshuo Yang and Weicai Ye and He Tong and Jingwen He and Yu Qiao and Hongsheng Li},
      year={2024},
      eprint={2405.05945},
      archivePrefix={arXiv},
      primaryClass={cs.CV},
      url={https://arxiv.org/abs/2405.05945}, 
}

@article{li2024hunyuan,
  title={Hunyuan-dit: A powerful multi-resolution diffusion transformer with fine-grained chinese understanding},
  author={Li, Zhimin and Zhang, Jianwei and Lin, Qin and Xiong, Jiangfeng and Long, Yanxin and Deng, Xinchi and Zhang, Yingfang and Liu, Xingchao and Huang, Minbin and Xiao, Zedong and others},
  journal={arXiv preprint arXiv:2405.08748},
  year={2024}
}

@inproceedings{peebles2023scalable,
  title={Scalable diffusion models with transformers},
  author={Peebles, William and Xie, Saining},
  booktitle={Proceedings of the IEEE/CVF international conference on computer vision},
  pages={4195--4205},
  year={2023}
}

@article{podell2023sdxl,
  title={Sdxl: Improving latent diffusion models for high-resolution image synthesis},
  author={Podell, Dustin and English, Zion and Lacey, Kyle and Blattmann, Andreas and Dockhorn, Tim and M{\"u}ller, Jonas and Penna, Joe and Rombach, Robin},
  journal={arXiv preprint arXiv:2307.01952},
  year={2023}
}

@inproceedings{esser2024scaling,
  title={Scaling rectified flow transformers for high-resolution image synthesis},
  author={Esser, Patrick and Kulal, Sumith and Blattmann, Andreas and Entezari, Rahim and M{\"u}ller, Jonas and Saini, Harry and Levi, Yam and Lorenz, Dominik and Sauer, Axel and Boesel, Frederic and others},
  booktitle={Forty-first international conference on machine learning},
  year={2024}
}

@inproceedings{tang2025aerogen,
  title={AeroGen: Enhancing remote sensing object detection with diffusion-driven data generation},
  author={Tang, Datao and Cao, Xiangyong and Wu, Xuan and Li, Jialin and Yao, Jing and Bai, Xueru and Jiang, Dongsheng and Li, Yin and Meng, Deyu},
  booktitle={Proceedings of the Computer Vision and Pattern Recognition Conference},
  pages={3614--3624},
  year={2025}
}

@article{tang2025terragen,
  title={TerraGen: A Unified Multi-Task Layout Generation Framework for Remote Sensing Data Augmentation},
  author={Tang, Datao and Wang, Hao and Xin, Yudeng and Qiao, Hui and Jiang, Dongsheng and Li, Yin and Yu, Zhiheng and Cao, Xiangyong},
  journal={arXiv preprint arXiv:2510.21391},
  year={2025}
}

@article{ye2025object,
  title={Object Fidelity Diffusion for Remote Sensing Image Generation},
  author={Ye, Ziqi and Ma, Shuran and Yang, Jie and Yang, Xiaoyi and Gong, Ziyang and Yang, Xue and Wang, Haipeng},
  journal={arXiv preprint arXiv:2508.10801},
  year={2025}
}

@inproceedings{zhang2023adding,
  title={Adding conditional control to text-to-image diffusion models},
  author={Zhang, Lvmin and Rao, Anyi and Agrawala, Maneesh},
  booktitle={Proceedings of the IEEE/CVF international conference on computer vision},
  pages={3836--3847},
  year={2023}
}

@inproceedings{li2024controlnet++,
  title={Controlnet++: Improving conditional controls with efficient consistency feedback: Project page: liming-ai. github. io/controlnet\_plus\_plus},
  author={Li, Ming and Yang, Taojiannan and Kuang, Huafeng and Wu, Jie and Wang, Zhaoning and Xiao, Xuefeng and Chen, Chen},
  booktitle={European Conference on Computer Vision},
  pages={129--147},
  year={2024},
  organization={Springer}
}

@article{zhao2023uni,
  title={Uni-controlnet: All-in-one control to text-to-image diffusion models},
  author={Zhao, Shihao and Chen, Dongdong and Chen, Yen-Chun and Bao, Jianmin and Hao, Shaozhe and Yuan, Lu and Wong, Kwan-Yee K},
  journal={Advances in Neural Information Processing Systems},
  volume={36},
  pages={11127--11150},
  year={2023}
}

@article{peng2024controlnext,
  title={Controlnext: Powerful and efficient control for image and video generation},
  author={Peng, Bohao and Wang, Jian and Zhang, Yuechen and Li, Wenbo and Yang, Ming-Chang and Jia, Jiaya},
  journal={arXiv preprint arXiv:2408.06070},
  year={2024}
}

@misc{
  stabilityai_sd35_release_2024,
  title={{Introducing Stable Diffusion 3.5}},
  author={{Stability AI}},
  year={2024},
}

@inproceedings{zhang2025creatilayout,
  title={Creatilayout: Siamese multimodal diffusion transformer for creative layout-to-image generation},
  author={Zhang, Hui and Hong, Dexiang and Wang, Yitong and Shao, Jie and Wu, Xinglong and Wu, Zuxuan and Jiang, Yu-Gang},
  booktitle={Proceedings of the IEEE/CVF International Conference on Computer Vision},
  pages={18487--18497},
  year={2025}
}

@article{beaumont2022laion,
  title={LAION-5B: A new era of open large-scale multi-modal datasets},
  author={Beaumont, Romain},
  journal={Laion. ai},
  year={2022}
}

@article{gadre2023datacomp,
  title={Datacomp: In search of the next generation of multimodal datasets},
  author={Gadre, Samir Yitzhak and Ilharco, Gabriel and Fang, Alex and Hayase, Jonathan and Smyrnis, Georgios and Nguyen, Thao and Marten, Ryan and Wortsman, Mitchell and Ghosh, Dhruba and Zhang, Jieyu and others},
  journal={Advances in Neural Information Processing Systems},
  volume={36},
  pages={27092--27112},
  year={2023}
}

@article{yuan2024fusu,
  title={FUSU: A multi-temporal-source land use change segmentation dataset for fine-grained urban semantic understanding},
  author={Yuan, Shuai and Lin, Guancong and Zhang, Lixian and Dong, Runmin and Zhang, Jinxiao and Chen, Shuang and Zheng, Juepeng and Wang, Jie and Fu, Haohuan},
  journal={Advances in Neural Information Processing Systems},
  volume={37},
  pages={132417--132439},
  year={2024}
}

@article{wang2021loveda,
  title={LoveDA: A remote sensing land-cover dataset for domain adaptive semantic segmentation},
  author={Wang, Junjue and Zheng, Zhuo and Ma, Ailong and Lu, Xiaoyan and Zhong, Yanfei},
  journal={arXiv preprint arXiv:2110.08733},
  year={2021}
}

@article{wang2025unicombine,
  title={Unicombine: Unified multi-conditional combination with diffusion transformer},
  author={Wang, Haoxuan and Peng, Jinlong and He, Qingdong and Yang, Hao and Jin, Ying and Wu, Jiafu and Hu, Xiaobin and Pan, Yanjie and Gan, Zhenye and Chi, Mingmin and others},
  journal={arXiv preprint arXiv:2503.09277},
  year={2025}
}

@inproceedings{mou2024t2i,
  title={T2i-adapter: Learning adapters to dig out more controllable ability for text-to-image diffusion models},
  author={Mou, Chong and Wang, Xintao and Xie, Liangbin and Wu, Yanze and Zhang, Jian and Qi, Zhongang and Shan, Ying},
  booktitle={Proceedings of the AAAI conference on artificial intelligence},
  volume={38},
  number={5},
  pages={4296--4304},
  year={2024}
}

@article{wu2025less,
  title={Less-to-more generalization: Unlocking more controllability by in-context generation},
  author={Wu, Shaojin and Huang, Mengqi and Wu, Wenxu and Cheng, Yufeng and Ding, Fei and He, Qian},
  journal={arXiv preprint arXiv:2504.02160},
  year={2025}
}

@inproceedings{tan2025ominicontrol,
  title={Ominicontrol: Minimal and universal control for diffusion transformer},
  author={Tan, Zhenxiong and Liu, Songhua and Yang, Xingyi and Xue, Qiaochu and Wang, Xinchao},
  booktitle={Proceedings of the IEEE/CVF International Conference on Computer Vision},
  pages={14940--14950},
  year={2025}
}

@inproceedings{zhang2025easycontrol,
  title={Easycontrol: Adding efficient and flexible control for diffusion transformer},
  author={Zhang, Yuxuan and Yuan, Yirui and Song, Yiren and Wang, Haofan and Liu, Jiaming},
  booktitle={Proceedings of the IEEE/CVF International Conference on Computer Vision},
  pages={19513--19524},
  year={2025}
}

@inproceedings{chen2019learning,
  title={Learning semantic segmentation from synthetic data: A geometrically guided input-output adaptation approach},
  author={Chen, Yuhua and Li, Wen and Chen, Xiaoran and Gool, Luc Van},
  booktitle={Proceedings of the IEEE/CVF conference on computer vision and pattern recognition},
  pages={1841--1850},
  year={2019}
}

@inproceedings{toker2024satsynth,
  title={Satsynth: Augmenting image-mask pairs through diffusion models for aerial semantic segmentation},
  author={Toker, Aysim and Eisenberger, Marvin and Cremers, Daniel and Leal-Taix{\'e}, Laura},
  booktitle={Proceedings of the IEEE/CVF Conference on Computer Vision and Pattern Recognition},
  pages={27695--27705},
  year={2024}
}

@article{khanna2023diffusionsat,
  title={Diffusionsat: A generative foundation model for satellite imagery},
  author={Khanna, Samar and Liu, Patrick and Zhou, Linqi and Meng, Chenlin and Rombach, Robin and Burke, Marshall and Lobell, David and Ermon, Stefano},
  journal={arXiv preprint arXiv:2312.03606},
  year={2023}
}

@misc{
  stable-diffusion-v1-5,
  author = {
    Stability AI and RunwayML and LAION
  },
  title = {{Stable Diffusion v1-5 Model Card}},
  year = {2022}
}

@inproceedings{wang2025dataset,
  title={Dataset distillation with neural characteristic function: A minmax perspective},
  author={Wang, Shaobo and Yang, Yicun and Liu, Zhiyuan and Sun, Chenghao and Hu, Xuming and He, Conghui and Zhang, Linfeng},
  booktitle={Proceedings of the Computer Vision and Pattern Recognition Conference},
  pages={25570--25580},
  year={2025}
}

@inproceedings{sajedi2023datadam,
  title={Datadam: Efficient dataset distillation with attention matching},
  author={Sajedi, Ahmad and Khaki, Samir and Amjadian, Ehsan and Liu, Lucy Z and Lawryshyn, Yuri A and Plataniotis, Konstantinos N},
  booktitle={Proceedings of the IEEE/CVF International Conference on Computer Vision},
  pages={17097--17107},
  year={2023}
}

@inproceedings{cui2023scaling,
  title={Scaling up dataset distillation to imagenet-1k with constant memory},
  author={Cui, Justin and Wang, Ruochen and Si, Si and Hsieh, Cho-Jui},
  booktitle={International Conference on Machine Learning},
  pages={6565--6590},
  year={2023},
  organization={PMLR}
}

@inproceedings{zheng2023layoutdiffusion,
  title={Layoutdiffusion: Controllable diffusion model for layout-to-image generation},
  author={Zheng, Guangcong and Zhou, Xianpan and Li, Xuewei and Qi, Zhongang and Shan, Ying and Li, Xi},
  booktitle={Proceedings of the IEEE/CVF Conference on Computer Vision and Pattern Recognition},
  pages={22490--22499},
  year={2023}
}

@inproceedings{du2023minimizing,
  title={Minimizing the accumulated trajectory error to improve dataset distillation},
  author={Du, Jiawei and Jiang, Yidi and Tan, Vincent YF and Zhou, Joey Tianyi and Li, Haizhou},
  booktitle={Proceedings of the IEEE/CVF conference on computer vision and pattern recognition},
  pages={3749--3758},
  year={2023}
}

@inproceedings{cazenavette2022dataset,
  title={Dataset distillation by matching training trajectories},
  author={Cazenavette, George and Wang, Tongzhou and Torralba, Antonio and Efros, Alexei A and Zhu, Jun-Yan},
  booktitle={Proceedings of the IEEE/CVF Conference on Computer Vision and Pattern Recognition},
  pages={4750--4759},
  year={2022}
}

@article{lipman2022flow,
  title={Flow matching for generative modeling},
  author={Lipman, Yaron and Chen, Ricky TQ and Ben-Hamu, Heli and Nickel, Maximilian and Le, Matt},
  journal={arXiv preprint arXiv:2210.02747},
  year={2022}
}

@article{liu2022flow,
  title={Flow straight and fast: Learning to generate and transfer data with rectified flow},
  author={Liu, Xingchao and Gong, Chengyue and Liu, Qiang},
  journal={arXiv preprint arXiv:2209.03003},
  year={2022}
}

@inproceedings{kim2022dataset,
  title={Dataset condensation via efficient synthetic-data parameterization},
  author={Kim, Jang-Hyun and Kim, Jinuk and Oh, Seong Joon and Yun, Sangdoo and Song, Hwanjun and Jeong, Joonhyun and Ha, Jung-Woo and Song, Hyun Oh},
  booktitle={International Conference on Machine Learning},
  pages={11102--11118},
  year={2022},
  organization={PMLR}
}

@inproceedings{lee2022dataset,
  title={Dataset condensation with contrastive signals},
  author={Lee, Saehyung and Chun, Sanghyuk and Jung, Sangwon and Yun, Sangdoo and Yoon, Sungroh},
  booktitle={International Conference on Machine Learning},
  pages={12352--12364},
  year={2022},
  organization={PMLR}
}

@misc{mmseg2020,
    title = {{MMSegmentation} - {OpenMMLab} {Semantic} {Segmentation} {Toolbox} and {Benchmark}},
    author = {{MMSegmentation Contributors} and {open-mmlab}},
    month = jul,
    year = {2020},
    howpublished = {\url{https://github.com/open-mmlab/mmsegmentation}}
}

@inproceedings{xue2023freestyle,
  title={Freestyle layout-to-image synthesis},
  author={Xue, Han and Huang, Zhiwu and Sun, Qianru and Song, Li and Zhang, Wenjun},
  booktitle={Proceedings of the IEEE/CVF conference on computer vision and pattern recognition},
  pages={14256--14266},
  year={2023}
}

@inproceedings{zhao2023label,
  title={Label freedom: Stable diffusion for remote sensing image semantic segmentation data generation},
  author={Zhao, Chenbo and Ogawa, Yoshiki and Chen, Shenglong and Yang, Zhehui and Sekimoto, Yoshihide},
  booktitle={2023 IEEE International Conference on Big Data (BigData)},
  pages={1022--1030},
  year={2023},
  organization={IEEE}
}

@inproceedings{zhao2017pyramid,
  title={Pyramid scene parsing network},
  author={Zhao, Hengshuang and Shi, Jianping and Qi, Xiaojuan and Wang, Xiaogang and Jia, Jiaya},
  booktitle={Proceedings of the IEEE conference on computer vision and pattern recognition},
  pages={2881--2890},
  year={2017}
}

@inproceedings{kulikov2025flowedit,
  title={Flowedit: Inversion-free text-based editing using pre-trained flow models},
  author={Kulikov, Vladimir and Kleiner, Matan and Huberman-Spiegelglas, Inbar and Michaeli, Tomer},
  booktitle={Proceedings of the IEEE/CVF International Conference on Computer Vision},
  pages={19721--19730},
  year={2025}
}

@article{yoon2025splitflow,
  title={SplitFlow: Flow Decomposition for Inversion-Free Text-to-Image Editing},
  author={Yoon, Sung-Hoon and Li, Minghan and Beaudouin, Gaspard and Wen, Congcong and Azhar, Muhammad Rafay and Wang, Mengyu},
  journal={arXiv preprint arXiv:2510.25970},
  year={2025}
}

@inproceedings{strudel2021segmenter,
  title={Segmenter: Transformer for semantic segmentation},
  author={Strudel, Robin and Garcia, Ricardo and Laptev, Ivan and Schmid, Cordelia},
  booktitle={Proceedings of the IEEE/CVF international conference on computer vision},
  pages={7262--7272},
  year={2021}
}

@inproceedings{cheng2022masked,
  title={Masked-attention mask transformer for universal image segmentation},
  author={Cheng, Bowen and Misra, Ishan and Schwing, Alexander G and Kirillov, Alexander and Girdhar, Rohit},
  booktitle={Proceedings of the IEEE/CVF conference on computer vision and pattern recognition},
  pages={1290--1299},
  year={2022}
}
}
\clearpage
\setcounter{page}{1}
\appendix
\maketitlesupplementary


\begin{figure*}[h!]
 \centering
 \includegraphics[width=1.0\linewidth]{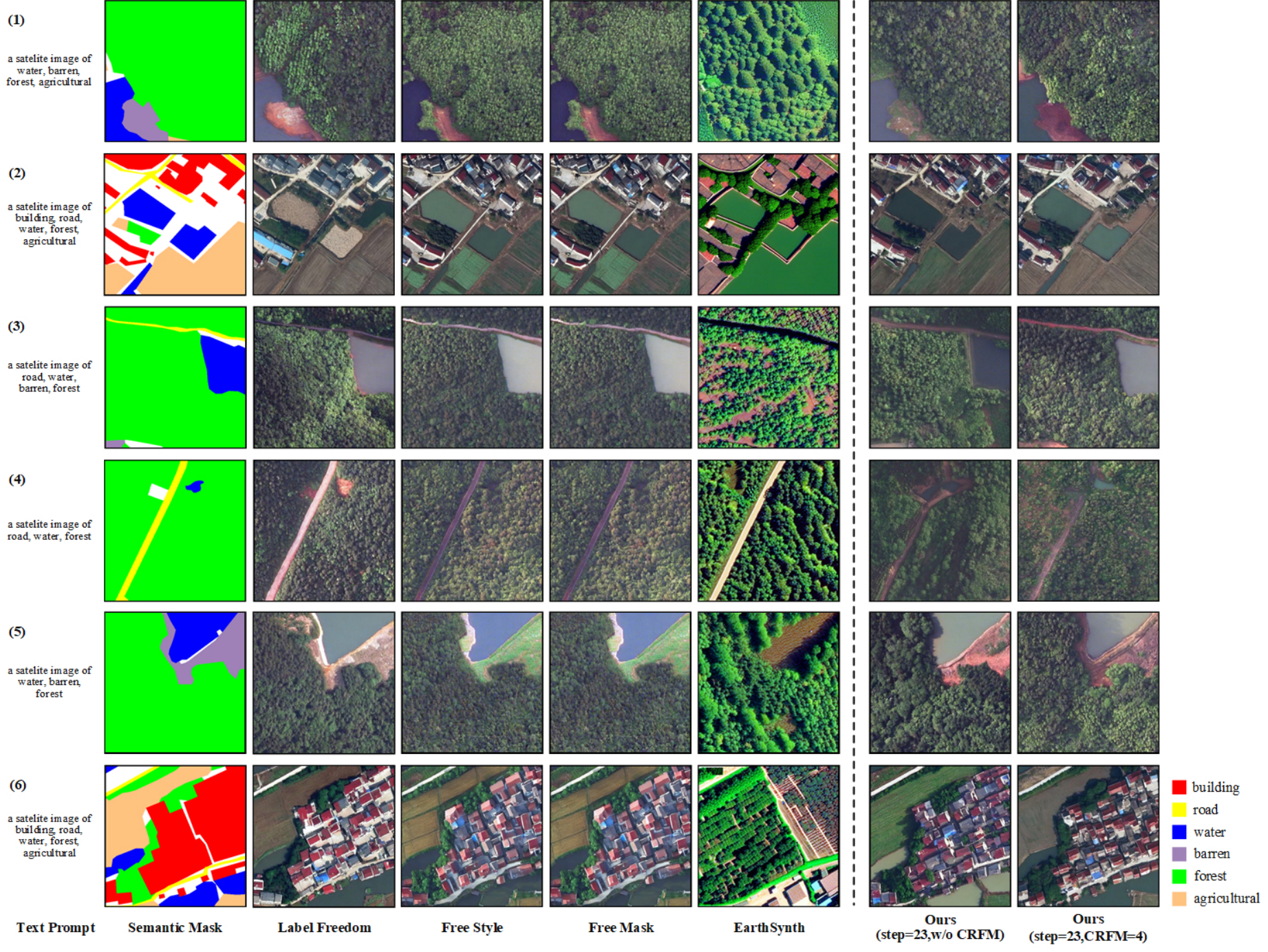}
 \caption{Qualitative comparison for mask-to-image generation by various methods.}
 \label{fig:S1}
\end{figure*}

\section{Differences Between CRFM and Inversion-Free Editing}
\label{sec:discussions}

Alg.~\ref{alg:crfm} shows the pseudo-code of Control-Rectify Flow Matching (CRFM), which is designed to modulate the trajectory during the early RF process, guiding it toward the target conditional distribution and thereby improving the semantic alignment of the final generated images.

The concept of steering the rectified flow (RF) trajectory has also been similarly applied in image editing tasks. FlowEdit~\cite{kulikov2025flowedit} modulates the trajectory of the RF by calculating the vector field difference between the source and target images at each time step. This approach obviates the need for latent inversion, and directly establishes the transition trajectory between the source and target images for image editing tasks. SplitFlow~\cite{yoon2025splitflow} further introduces a flow decomposition and aggregation framework, designed to address the inherent issues of gradient entanglement and semantic conflict within complex target prompts, which mitigates interference between attributes while maintaining overall semantic consistency.

The trajectories established by the aforementioned two methods represent a transformation process from the source data distribution $\boldsymbol{p}_{\text{data}}^{\text{src}}$ to the target data distribution $\boldsymbol{p}_{\text{data}}^{\text{tgt}}$. This process can be formally expressed in the form of the following ODE:
\begin{equation}
dz^{edit}=v_t^{\Delta}(z_t^{src},z_t^{tgt})dt,
\end{equation}
where $z^{edit}$ denotes the latent state on the editing trajectory, while $z_t^{src}$ and $z_t^{tgt}$ represent the interpolated latent representations of the source and target states, respectively, at time $t \in [0, 1]$. Specifically, the velocity field difference $v_t^{\Delta}(z_t^{src},z_t^{tgt})$ is defined as the subtraction of the source velocity field from the target velocity field, where $v(z_t,t)$ is the predicted velocity at latent state $z_t $ and time $t$:
\begin{equation}
v_t^{\Delta}(z_t^{src},z_t^{tgt}) = v(z_t^{tgt},t) - v(z_t^{src},t).
\end{equation}

\begin{algorithm}[ht]
\caption{Control-Rectify Flow Matching}
\label{alg:crfm}
\SetKwInOut{Input}{Input}
\SetKwInOut{Output}{Output}

\Input{Vector field predictions $v_\theta$, Timestep latents $z_t$, Noise schedule $\sigma_t$, VAE Decoder $\mathcal{D}$, Conditional model $\Phi$, Ground truth labels $Y_{gt}$}
\Output{Rectified vector field $v^*$}

\BlankLine
$z_0' \leftarrow z_t - \sigma_t \cdot v_\theta$ \;
\BlankLine
$x_0' \leftarrow \mathcal{D}(z_0')$
\BlankLine
$\hat{Y} \leftarrow \Phi(x_0')$ \;
\BlankLine
$\mathcal{L}_{cond} \leftarrow \text{CrossEntropy}(\hat{Y}, Y_{gt})$ \;
\BlankLine
$g \leftarrow \nabla_{v_\theta} \mathcal{L}_{cond}$ \;
\BlankLine
$v^* \leftarrow v_\theta - g$ \;
\BlankLine
\Return{$v^*$}
\end{algorithm}

In contrast, our CRFM modulates the trajectory that connects the noise distribution $z_1 \sim \mathcal{N}(0, 1)$ to the conditional data distribution $z_0 \sim p(z_{data} | C^m)$ which is controlled by a semantic mask $C^m$.The probability path of this process can be constructed by solving the following ODE:
\begin{equation}
dz^{ctrl}=v_t^{CRFM}(z_t,t,C^m)dt,
\end{equation}
where $dz^{ctrl}$ is the latent state on the conditional trajectory. The term $v_t^{CRFM}(z_t,t,C^{m})$ is the vector field adjusted by CRFM based on the semantic mask condition $C^{m}$ at the current latent state $z_t$ (at time $t$). CRFM vector field can be decomposed into two distinct components:
\begin{equation}
v_t^{CRFM}(z_t,t,C^{m})=v_t(z_t,t,C^{m})+v_t^{rec}(z_t,t,C^{m}).
\end{equation}
Here, $v_t(z_t,t,C^m)$ represents the output vector field of RF model, conditioned on the semantic mask $C^{m}$ and computed at the current latent state $z_t$ and time $t$. $v_t^{rec}(z_t,t,C^{m})$, is an estimated vector derived from the gradient, calculated using a Segmentation Network $\mathcal{S}$ and the semantic mask $\boldsymbol{C}^{\boldsymbol{m}}$ on the pre-synthesized image $z_0^t$ of $z_t$. This component is formally defined as:
\begin{equation}
v_t^{rec}(z_t,t,C^{m})\approx\alpha\nabla_{v}\mathcal{L}_{CE}(\mathcal{S}(z_0^t), C^m),
\end{equation}
\begin{equation}
z_0^t = z_t - \sigma_t v_t(z_t, t, C^m),
\end{equation}
where $\alpha$ is a manually set hyperparameter, $\sigma_t$ is the schedule coefficient that governs the blend ratio between the noise and data during the forward process of RF. Furthermore, $\mathcal{L}_{CE}(\mathcal{S}(z_0^t), C^m)$ calculates the semantic loss between the output of the Segmentation Network $\mathcal{S}(z_0^t)$ and the semantic mask $C^{m}$.

Beyond these technical differences in trajectory construction, the core motivations of the two approaches are fundamentally distinct. FlowEdit and related editing methods are built on the principle that an effective editing algorithm should minimally alter the source image while faithfully transporting it toward the target distribution. Their objective is to construct a minimal and direct transition between $p_{data}^{src}$ to $p_{data}^{tgt}$.


In sharp contrast, CRFM aims to modulate the probability trajectory during the RF process for any initial state $z_1 \sim \mathcal{N}(0, 1)$, ensuring that the resulting generated state $z_0$ more closely aligns with the target data distribution $p(z_{data}|C^{m})$.

\section{Additional Experiment}
\label{sec:experiments}

\begin{table*}[h]
    \centering
    \caption{The performance of downstream tasks with varying pre-trained semantic segmentation models on CRFM.}
    \label{tab:S1}
    \begin{tabular}{l|*{3}{ccc}}
        \toprule
        {\textbf{Method}} & \multicolumn{3}{c}{Seg. Model-1} & \multicolumn{3}{c}{Seg. Model-2} & \multicolumn{3}{c}{Seg. Model-3} \\
        \cmidrule(lr){2-4} \cmidrule(lr){5-7} \cmidrule(lr){8-10}
        & \textbf{OA} & \textbf{mIoU} & \textbf{mAcc} & \textbf{OA} & \textbf{mIoU} & \textbf{mAcc} & \textbf{OA} & \textbf{mIoU} & \textbf{mAcc} \\
        \midrule
        Pre-trained Seg. model (baseline) & 74.27 & 45.27 
        & 56.44 & 69.11 & 40.76 & 52.98 & 67.98 & 30.12 & 40.48 \\ \hline
        Downstream task(step=18, CRFM=2) & 75.86 & 47.99 & 59.46 & 75.11 & 47.30 & 59.89 & 75.07 & 46.35 & 58.64 \\
        Downstream task(step=18, CRFM=4) & 76.24 & 48.93 & 60.54 & 75.46 & 48.35 & 60.82 & 75.47 & 47.65 & 60.19 \\
        Downstream task(step=18, CRFM=6) & \textbf{76.02} & \textbf{49.57} & \textbf{62.19} & \textbf{75.66} & \textbf{48.92} & \textbf{61.14} & \textbf{75.58} & \textbf{48.44} & \textbf{60.67} \\
        \bottomrule
    \end{tabular}
\end{table*}

\begin{figure*}[h!]
 \centering
 \includegraphics[width=1.0\linewidth]{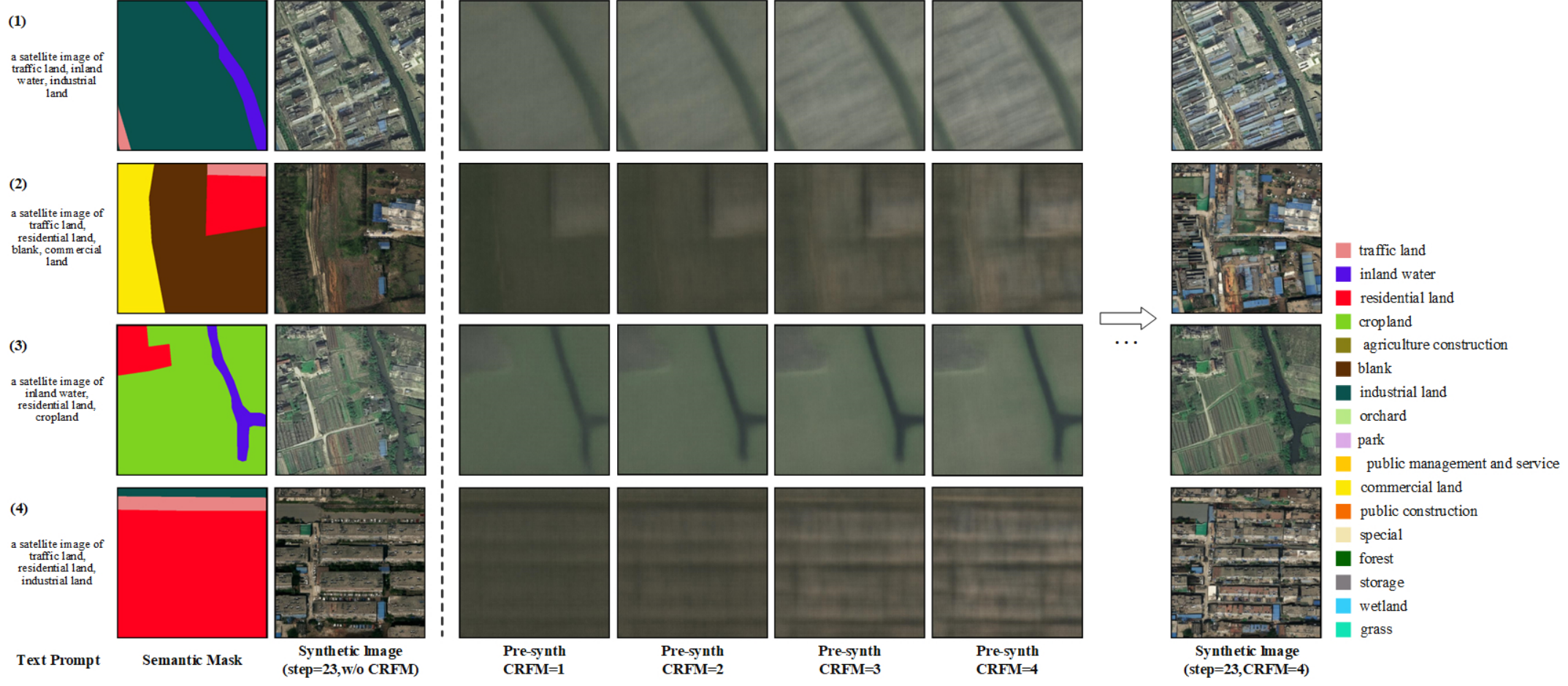}
 \caption{Visualization results of pre-synth images with different CRFM steps on the FUSU.}
 \label{fig:S2}
\end{figure*}

\begin{figure*}[h!]
 \centering
 \includegraphics[width=1.0\linewidth]{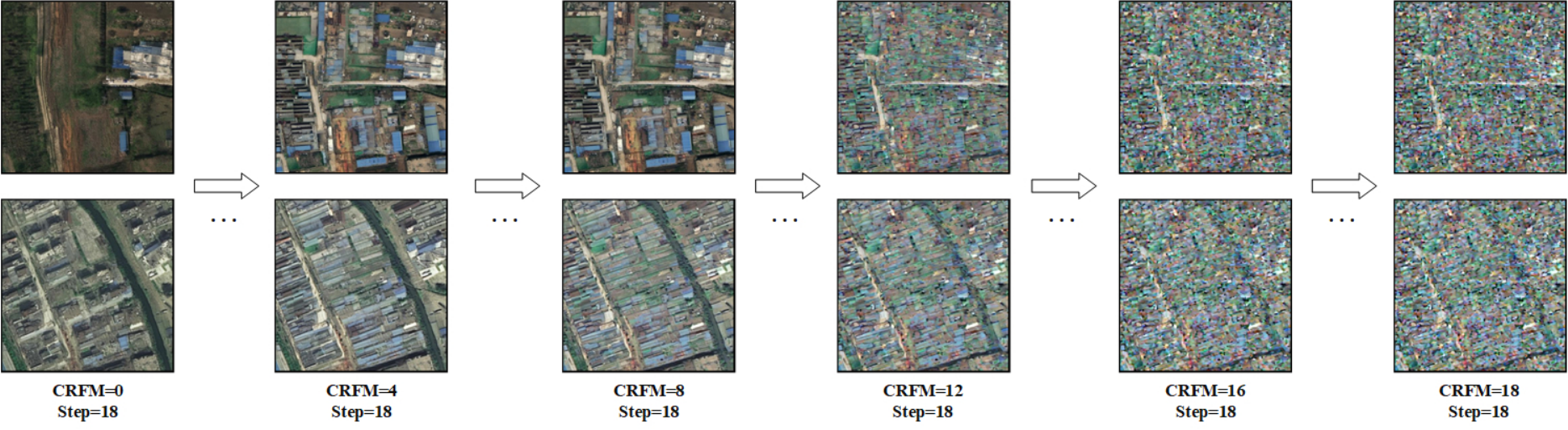}
 \caption{Visual comparison of generated samples with varying CRFM steps.}
 \label{fig:S3}
\end{figure*}

\subsection{Visualization of Comparative Results}

We provide the visualization results of different methods on the LoveDA~\cite{wang2021loveda} dataset, including LabelFreedom~\cite{zhao2023label}, Freestyle~\cite{xue2023freestyle}, FreeMask~\cite{yang2023freemask}, and EarthSynth~\cite{pan2025earthsynth}, all of which are based on ControlNet~\cite{zhang2023adding}. As shown in Figure~\ref{fig:S1}, the ControlNet architecture achieves stronger edge control because the feature outputs from its semantic mask control flow are directly added into the intermediate layers of the image diffusion branch via skip connections. This localized injection mechanism provides tight spatial conditioning.

In contrast, the MMDiT architecture processes the concatenated features of text and the semantic mask through a global Transformer-based attention mechanism. This architectural difference enables MMDiT to model long-range dependencies and better capture global context, thereby resulting in more natural and aesthetically pleasing generation and higher semantic consistency in complex scenes. By optimizing the vector field's direction in the pixel-level latent space, CRFM significantly enhances the precision of edge control, thereby complementing the shortcomings of MMDiT in this regard.

\begin{table*}[t]
    \caption{The improvement of Control-Rectify Flow Matching on rare categories.}
    \label{tab:S2}
  \setlength{\tabcolsep}{12pt}
  \centering
\begin{tabular}{lcccccc}
\toprule
\textbf{Method} & \multicolumn{2}{c}{\textbf{Orchard}} & \multicolumn{2}{c}{\textbf{Pub. Mngmt.}} & \multicolumn{2}{c}{\textbf{Storage}} \\
\cmidrule(lr){2-3} \cmidrule(lr){4-5} \cmidrule(lr){6-7}
 & \textbf{IoU} & \textbf{Acc} & \textbf{IoU} & \textbf{Acc} & \textbf{IoU} & \textbf{Acc} \\ \midrule
Baseline~\cite{zhao2017pyramid} & 30.01 & 40.65 & 39.36 & 48.60 & 32.72 & 39.33 \\
FreeMask~\cite{yang2023freemask} & 37.71 & 42.90 & 41.62 & 54.50 & 32.60 & 43.95 \\
SD v3.5~\cite{esser2024scaling} & 37.73 & 45.26 & 41.58 & 54.45 & 33.86 & 49.54 \\
Ours & \textbf{41.88} & \textbf{55.37} & \textbf{42.79} & \textbf{58.00} & \textbf{40.10} & \textbf{56.91}  \\ \bottomrule
\end{tabular}
\vspace{-3mm}
\end{table*}


\subsection{Sensitivity Analysis of CRFM on Pre-Trained Semantic Segmentation Models}

For the semantic segmentation model, we employ PSPNet~\cite{zhao2017pyramid}. This choice is supported by the comparative analysis in TISynth~\cite{DONG2026103839}, which evaluates architectures including PSPNet, Segmenter~\cite{strudel2021segmenter}, and Mask2Former~\cite{cheng2022masked}. Their findings indicates that CNN-based architectures demonstrate a superior performance advantage over transformer-based architectures for small datasets like FUSU-4k~\cite{yuan2024fusu}.

Because CRFM relies on the semantic segmentation model to compute the semantic loss between its prediction and the ground truth mask, the quality of the pre-trained model may directly affects CRFM performance. To varify the sensitivity of CRFM on pre-trained semantic segmentation models, we design three segmentation models with different accuracies: Seg. Model-1 is trained on the full FUSU-4k dataset. Seg. Model-2 is trained on a 2k subset and Seg. Model-3 is trained on a 1k subset.

As shown in Table~\ref{tab:S1}, we first report the performance of the pre-trained semantic segmentation models. Using these models as the evaluation modules within CRFM, we compare CRFM under different number of adjustment steps (i.e.,2, 4 and 6), while fixing the sampling step at 18. When utilizing lower-performing pre-trained models, the effectiveness of CRFM is slightly reduced. However, as the number of adjustment steps increases, the final performance consistently surpasses the baseline. This indicates that although the pre-trained semantic segmentation models vary in accuracy, they all possess a certain level of semantic discrimination capability. Such information is sufficiently reliable to guide and correct the sampling trajectory during the early sampling stage. Therefore, although CRFM is somewhat sensitive to the performance of the pre-trained segmentation model, this influence is limited because CRFM operates only in the early sampling stage. Consequently, it consistently delivers improvements over the baseline.


\subsection{Performance Gains on Rare-Classes}

To evaluate the robustness of CRFM, we analyse experiments on three rare classes: Orchard, Public Management (Pub. Mngmt.), and Storage. As shown in Table~\ref{tab:S2}, our method achieves substantial performance leaps, consistently outperforming state-of-the-art baselines like SD v3.5~\cite{esser2024scaling} and FreeMask~\cite{yang2023freemask}. Notably, in the Storage domain, CRFM improves IoU from 33.86\% to 40.10\% and significantly boosts Accuracy (Acc) from 49.54\% to 56.91\%.

These gains provide strong empirical evidence for our CRFM mechanism. In data-scarce scenarios, the vanilla vector field $v_\theta$ or noise $\epsilon_\theta$ often provides biased or blurry guidance due to the scarcity of samples. By introducing the rectification term $g = \nabla_{v_\theta} \mathcal{L}_{cond}$, CRFM actively steers the flow trajectories $v^*$ toward the precise semantic manifold. The marked improvement in Acc—reaching 58.00\% in Pub. Mngmt.—underscores that our CRFM effectively restores category boundaries and high-frequency details that are typically lost in standard flow-matching trajectories and noise-based diffusion processes.

\subsection{Analysis of Sampling with CRFM}

To better understand the effect of CRFM on the flow trajectory, we visualize the pre-synth images generated by the CRFM algorithm during sampling, with the total sampling steps set to 18 and the number of CRFM adjustments set to 4. As shown in Figure~\ref{fig:S2}, the generated pre-synth images increasingly align with the semantic structure of the ground-truth masks as the CRFM adjustments proceed. This indicates that CRFM is capable of adjusting the flow trajectory of the noise toward $p(z_{data}|C^m)$, a conclusion further supported by the comparison between generated images with and without CRFM.

However, we observe that setting the number of CRFM adjustments excessively high—typically exceeding $50\%$ of the total sampling steps leads to mode collapse. This phenomenon occurs because the generated images over-optimize to satisfy the pre-trained segmentation model's perception, thereby resulting in a noticeable degradation of human-perceived visual quality. Figure~\ref{fig:S3} presents the visualization results for a total of 18 sampling steps, while varying the CRFM adjustment count from 0 to 18. A slight mode collapse appears at CRFM = 8, becomes severe at CRFM = 12, and when the CRFM frequency approaches the full sampling length, the outputs degenerate into adversarial-like samples that completely lose visual quality.

\end{document}